# Post-disaster building indoor damage and survivor detection using autonomous path planning and deep learning with unmanned aerial vehicles

Xiao Pan[1, 2*], Sina Tavasoli[2], T. Y. Yang[2], Sina Poorghasem[3]

[1] Department of Civil and Environmental Engineering, The Hong Kong University of Science and Technology, Clear Water Bay, Hong Kong, cexiaopan@ust.hk
[2] Department of Civil Engineering, University of British Columbia, Vancouver, Canada
[3] Stevens Institute of Technology, Hoboken, New Jersey, USA

**ABSTRACT**

Rapid response to natural disasters such as earthquakes is a crucial element in ensuring the safety of civil infrastructures and minimizing casualties. Traditional manual inspection is labour-intensive, time-consuming, and can be dangerous for inspectors and rescue workers. This paper proposed an autonomous inspection approach for structural damage inspection and survivor detection in the post-disaster building indoor scenario, which incorporates an autonomous navigation method, deep learning-based damage and survivor detection method, and a customized low-cost micro aerial vehicle (MAV) with onboard sensors. Experimental studies in a pseudo-post-disaster office building have shown the proposed methodology can achieve high accuracy in structural damage inspection and survivor detection. Overall, the proposed inspection approach shows great potential to improve the efficiency of existing manual post-disaster building inspection.

## 1. Introduction

Traditional post-disaster inspection is done manually, where human inspectors and rescue workers are deployed to a site to conduct inspection surveys and identify potential survivors. The process is time-consuming and labour-intensive, highly dependent on the rescue workers' expertise. Meanwhile, it can be risky for humans in extreme events such as major earthquake aftershocks. To address this, many efforts have been dedicated to automating the inspection process. One of the promising research directions is through robotic inspection approaches such as unmanned ground vehicles (UGVs) [1-2]. UGVs are deployed in scenarios such as survivor tracking, where robots are called upon to work in unknown environments [3], create models of their surroundings and know precisely where they are in those environments [4-5]. However, UGVs have some disadvantages in the post-disaster environment. For instance, navigating through debris is challenging which can constrain the UGV movement and even cause its overturning.[6] Also, the field of view is rather limited at ground level. [7-8]. On the other hand, Unmanned Aerial Vehicles (UAVs) are a promising alternative and have applications in survivor identification [9-10] and structure detection [11-12]. Most investigations have deployed large, manually operated UAVs for outdoor inspection [13-14]. For instance, Zhuge et al. (2022) [15] demonstrated multi-UAV bridge deflection sensors for improved structure health monitoring. Meng et al. (2023) [16] proposed a UAV-based real-time automatic crack detection that is combined with classification, segmentation, and measurement algorithms. Despite the promising outcomes, these studies require manual control of the UAV, without further efforts in automating the UAVs.

Full autonomous UAVs are restricted by the need for bulky sensors, requiring larger UAVs with massive payloads which may not be a good fit for post-disaster conditions. This is because large-scale UAVs such as the DJI Matrice series cannot pass through narrow passages and openings in indoor scenarios. The small- to medium-scale UAVs in post-disaster scenarios have recently been explored. For example, Dowling et al.'s (2018) [17] attempted indoor autonomous mapping using Unmanned Recon and Safety Aircraft (URSA). A. Farooq et al.'s (2022) [18] proposed a rescue framework with depth cameras and autonomous mapping. These studies lacked real-world validation and compatibility with small MAVs. On the other hand, some other studies attempted grid-based path planning and dead reckoning strategies [19-21]. Despite the achievements, these studies were conducted in a semi-automated automated, where a human operator is involved with the UAV to correct the cumulative errors from the IMU sensors. Similarly, Pliakos et al. (2024) [22] created an autonomous MAV for indoor rescue that can float partially alone but needs human guidance when in a hazardous situation. Recently, Tavasoli et al. (2024) [23], proposed a framework for autonomous survivor localization and mapping in post-disaster environments, where a lightweight LiDAR and thermal camera are used for efficient mapping and survivor localization, respectively.

Recent advances in computer graphics and artificial intelligence offer promising data-driven solutions to process the image or point cloud data collected by mobile robots [24-25]. Novel image and point cloud processing algorithms have been developed to automate the inspection, monitoring, and assessment of reinforced concrete (RC) structures [26-27], steel structures [28-29], timber structures [30], structural joints [31-32], and retaining walls [33], etc, where the effectiveness of the image and point cloud techniques has been demonstrated.

In this paper, an autonomous inspection approach has been proposed for simultaneous structural damage inspection and survivor detection in the building's indoor scenario. The proposed inspection workflow consists of an autonomous path planning and navigation method, and advanced image processing methods, which enables an MAV to self-navigate and perform inspection tasks. A series of pseudo-post-disaster experiments have been conducted in office rooms for damage and survivor detection. For the proof-of-concept, a human worker and RC specimens are put into one of the office rooms at The University of British Columbia where the MAV is navigating and collecting data. The results of survivor detection show the efficacy of the proposed methodology in post-disaster scenarios. Using thermal cameras for this purpose, the model achieved mean Average Precision (mAP) values of 0.995 for mAP@0.5 and 0.838 for mAP@0.5:0.95, indicating robust performance in post-disaster survivor detection. Also, pretrained RC damage detector has been employed where the results have shown the damage of interest in the RC specimens can be successfully identified.

## 2. Methodologies

In the proposed autonomous inspection workflow, multiple novel algorithms have been developed and integrated for accurate damage inspection, survivor detection, and further implemented with the MAV using an autonomous path planning and navigation approach, as shown in *Figure 1*.

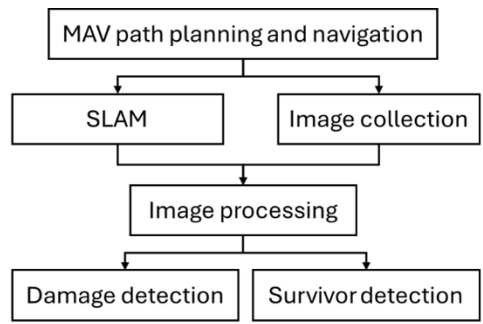

*Figure 1. Autonomous structural damage inspection and survivor detection workflow.*

### 2.1. Structural damage inspection

The proposed structural damage inspection involves novel image processing and 3D point cloud processing for damage localization and quantification.

2.1.1. Image-based damage detection

Image-based damage evaluation is used for preliminary assessment to identify the damage types and locations, while the 3D point cloud-based damage evaluation is implemented to provide a more comprehensive quantification of damage in 3D space. In this study, with reference to the ACI Guide for Conducting a Visual Inspection of Concrete in Service [ref], two of the most common damage types of reinforced concrete (RC) structures are selected, including concrete cracks and spalling, in order to demonstrate the proposed inspection approach. For this purpose, pretrained image-based RC damage detection algorithms are adopted [20].

2.1.2. 3D point cloud-based damage quantification

Apart from the image processing algorithms, a 3D point cloud algorithm is proposed to quantify structural damage in 3D space. First, a series of data preprocessing and filters are applied using the Open3D platform. Next, a point cloud segment-recovery strategy is adopted [24] to quantify the concrete spalling. Various types of point-cloud processing algorithms such as line fitting, arc fitting, and plane segmentation, can be considered for different types of structures such as steel or concrete components. *Figure 2* shows a rectangular RC column, where plane segmentation can be used to detect different sides of the column and recover the original undamaged configuration. The spalling depth and volume can then be quantified by slicing through a reference plane from the ground up.

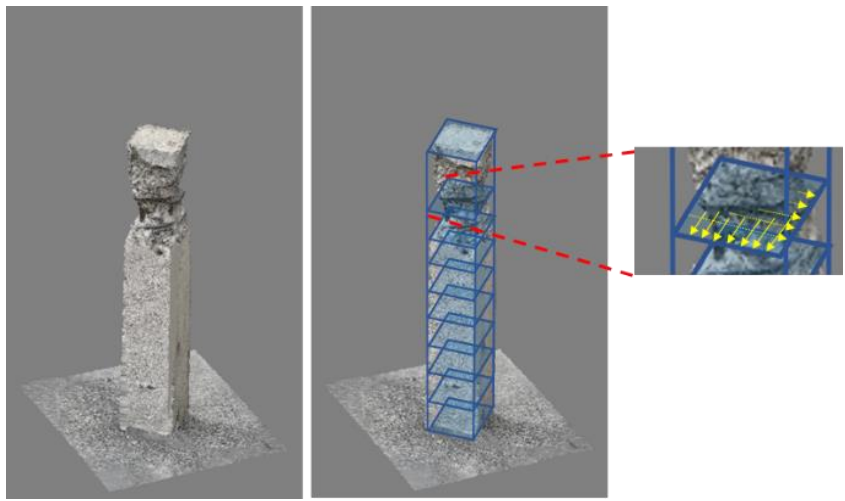

*Figure 2. Concrete damage quantification using 3D point cloud processing.*

**2.2. Thermal imaging–based survivor detection**

A specific variant of the You Only Look Once (YOLO) algorithms has been selected and integrated into a thermal-based survivor detection framework. The selected YOLOv8x delivers enhanced functionalities for object detection, image classification, and instance segmentation. This section describes the implementation of YOLOv8x on a custom dataset. The thermal images dataset was developed using the MLX90640 thermal Camera and the FLIR ONE Gen 3 iOS Thermal Camera. The MLX90640, featuring a 55°×35° field of view and a resolution of 32×24 pixels, was selected. The FLIR ONE thermal camera has a thermal resolution of 80×60 and is specifically designed to detect temperatures ranging from -4 to 248 degrees Fahrenheit, allowing for precise identification and documentation of heat radiation emitted by humans. This section focuses on detecting survivors in post-disaster scenarios such as earthquakes. Consequently, images were captured under various environmental conditions to simulate real-world disaster settings, as depicted in Figure 3. These images help train the model to differentiate between human heat signatures and other heat-emitting objects. Figure 3 (a) and (b) show exposed head and legs with the body obscured by debris, while Figure 3 (c) and (d) capture the survivor partially covered by fabrics and in low light conditions; and Figure 3 (e) and (f) illustrate a fully exposed body in normal and smoky conditions, respectively. The images were taken from approximately 1.5 meters above ground—the typical hovering elevation of a Micro Aerial Vehicle (MAV), which minimizes obstructions from surrounding structures. To enhance the algorithm's robustness, additional images at varying elevations and images without survivors were also incorporated into the dataset.

LabelMe software facilitated the manual labelling of humans in these images using axis-aligned bounding boxes, drawn around visible temperature gradients on the human body and labelled as 'human'. The bounding boxes were drawn with precision in size and position to accurately reflect visible human body parts. All images were resized to 640×640 pixels to align with the YOLOv8x model's input specifications. To expand the dataset's size and diversity, various image augmentation techniques were applied, including translation, rotation, flipping, scaling, cropping, brightness adjustments, and mosaic. This process involved randomly merging four images, followed by random modifications in translation, scaling, and rotation, and further augmented by flips and brightness adjustments. Finally, 2190 images were generated through these augmentation techniques and were randomly divided into training, validation, and testing sets in a 7:2:1 ratio.

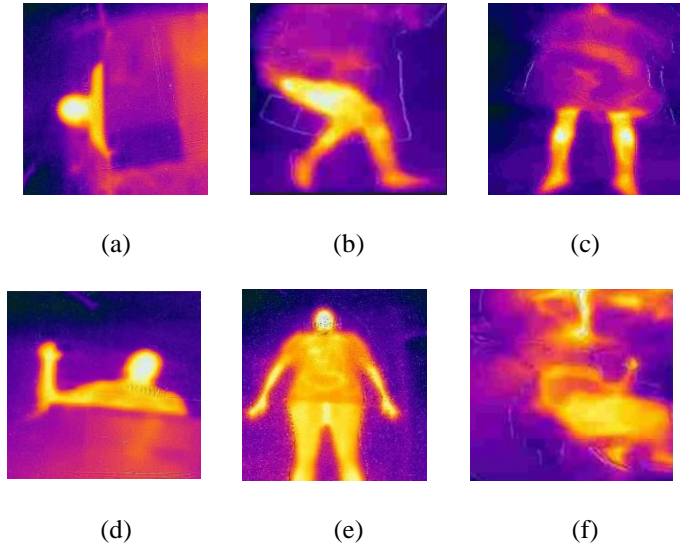

*Figure 3. Thermal images of a survivor in different conditions include: (a) head exposed while the body is under debris; (b) only legs visible while the body is covered by debris; (c) legs are visible while the head and body are under fabrics; (d) head and hand are visible while body is covered with debris in low light condition; (e) Entire of the survivor in standard condition; and (f) survivor in the presence of smog.*

**2.3. Autonomous path planning and navigation**

This study integrates frontier exploration [34] and the gmapping package developed in the Robot Operating System (ROS) to maximize the mapping effectiveness. This section elaborates on the autonomous path planning and navigation algorithm and the corresponding MAVs that are specialized for this task.

2.3.1. Description of the MAV and the sensor setup

The MAV used in this study is the Parrot Bebop 2. It has a 2000-metre range, 25 minutes of flight time, and a payload of 250g, making it one of the best programmable MAVs on the market for $300 USD. In contrast to other commercially available programmable MAVs, the Parrot Bebop 2 has a high payload and low price. The front-facing camera can capture 1080p Video or images at 4096 x 3072 resolution. The system also features an SF 45/B 2D LiDAR, one of the thinnest and lightest 2D LiDAR sensors on the market, at 49g with a horizontal viewing angle of 320° and a detection range of 50 meters. The internal processing is controlled by a Raspberry Pi 4 Model B Quad Core 64-Bit and offers high speed, multimedia, high memory, and connectivity. Power comes from the PiSugar 2 Pro Lithium Battery, a 5V 5000mAh battery capable of powering both the Raspberry Pi and the LiDAR. To detect survivors, a downward-facing MLX90640 thermal camera has been installed on the MAV's bottom.

2.3.2. Autonomous path planning and navigation algorithm

Frontier exploration focuses on marking the space between areas where the robot already knows and the ones that are unknown, allowing the robot to explore new territories instead of looking at random or thorough searches that often result in double visits. The pipeline also leverages the gazebo simulator, SLAM slam_gmapping, and amcl package for localization. The slam_gmapping node uses laser scan data to build a 2D map, showing open space, obstacles, and unexplored environments. The system uses the Adaptive Monte Carlo Localization (amcl) and the move_base nodes to localize and map paths for navigation. The amcl node uses particle filters to follow the position of the MAV, the global path planning is done by the rapidly-exploring random tree star (RRT*) algorithm and local planning is done by the dynamic window approach (DWA) algorithm. In contrast to conventional algorithms, such as A* or Dijkstra that apply in known and well-ordered environments, RRT* finds the path that is best in dynamic spaces, making it suitable for unstructured environments.

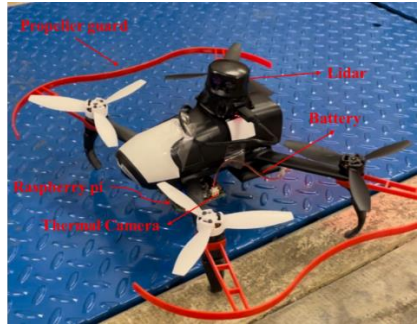

*Figure 4. MAV setup*

### 3. Experimental studies and results

#### 3.1. Description of the experimental setup

To demonstrate the proposed inspection approach, experimental studies have been conducted at The University of British Columbia. To simulate a pseudo-post-disaster scenario, damaged RC specimens were placed in multiple office rooms, and humans were lying on the ground with partially covered a body (*Figure 5*).

#### 3.2. Results and discussions

The MAV took off at a location near the entrance of the testing environment and utilized the packages and algorithms described in Section 2 to fly and collect data for navigation and path planning. In the experiment conducted, the flight height was set to 1.5 meters above the floor to approximate the mid-height of a typical building story. That height is selected so as to avoid interference from levels above and below. *Figure 6* demonstrates the actual floor plan and its projection onto the LiDAR-generated map of the MAV. The floor plan was 90 percent compatible with the mapping.

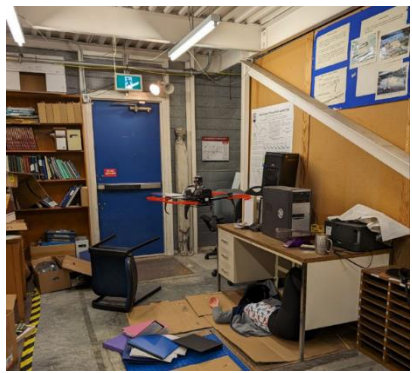

*Figure 5. Experimental office room*

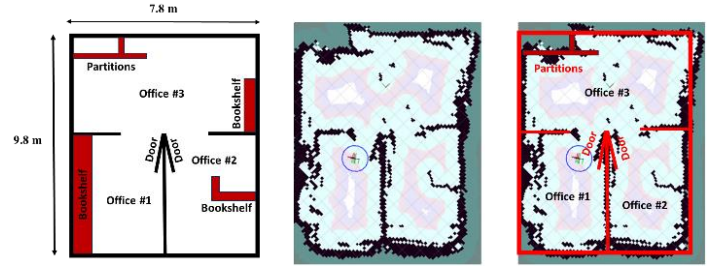

*Figure 6. SLAM results*

*Figure 7* shows the results of the damage detection on a concrete specimen. The RC columns and steel reinforcement exposure are successfully localized in all the images. In addition, 3D point clouds of RC columns are obtained using vision-based 3D reconstruction. Using 3D point cloud processing described in Section 2.1, the concrete spalling volume can be quantified. *Table 1* shows a summary of the concrete spalling quantification results. Besides, the estimation error with respect to the ground truth values is presented, which is around the range of 5-10% for the three columns examined. The average estimation error of the three columns is about 6.7%. Such errors may be attributed to the following aspects.

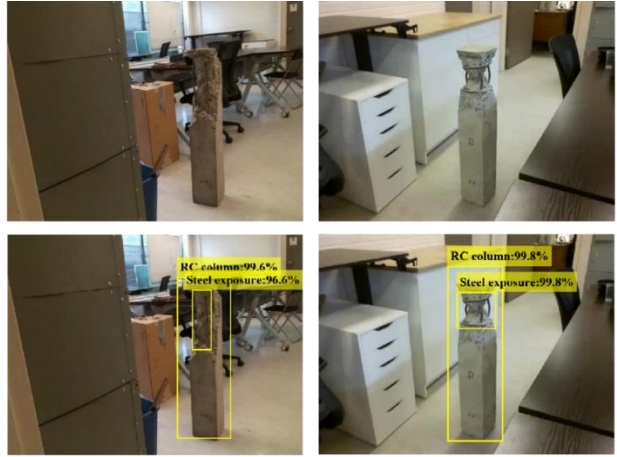

*Figure 7. Damage detection of RC specimens*

*Table 1. Spalling quantification of the RC columns*

|  | Estimated spalling volume [$cm^3$] | Ground truth [$cm^3$] | Error [%] |
|---|---|---|---|
| RC Column 1 | 5215 | 5868 | 11.1% |
| RC Column 2 | 12199 | 12956 | 5.8% |

For the thermal image-based survivor detection, the training of the YOLOv8x model was carried out using the PyTorch framework in Python. The model was trained for 800 epochs with a batch size of 16 and a learning rate of 0.01, employing a weight decay of 0.0005 with the AdamW optimizer. *Figure 8* illustrates the training and validation loss curves, demonstrating the model's converging performance. The effectiveness of the YOLOv8x model in survivor detection is evaluated using standard object detection metrics such as precision, recall, F1 score and mean Average Precision (mAP). These metrics, derived from both ground truth values and model predictions, provide a comprehensive evaluation of the model's accuracy in identifying survivors. Specifically, the model was evaluated using two mAP values, mAP@0.5 and

mAP@0.5:0.95, where mAP@0.5 is the value when the Intersection over Union (IoU) exceeds 0.5, and mAP@0.5:0.95 represents the average mAP calculated over an IoU range from 0.5 to 0.95, in increments of 0.05.

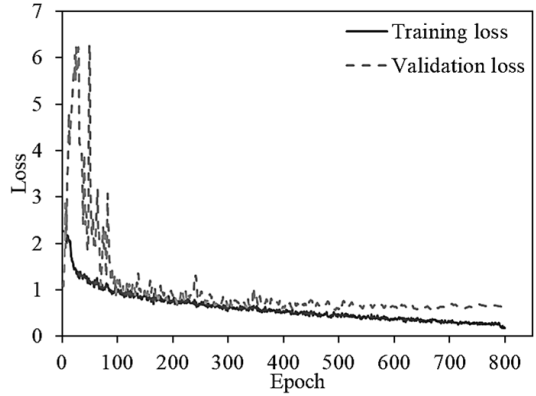

*Figure 8. Training and validation loss curves of YOLOv8x.*

The model reaches the precision, recall, and F1 score of 0.998, 1, and 0.99, respectively, on the training dataset. Additionally, the model achieved mAP@0.5 and mAP@0.5:0.95 scores of 0.995 and 0.838, respectively. This trained model demonstrated effective survivor detection in the test dataset, with true positive rates of 0.995 for mAP@0.5 and 0.837 for mAP@0.5:0.95. Model efficiency was further evaluated by an average processing time of less than 0.06 seconds per picture. *Figure 9* shows a close alignment between the prediction and the ground truth in both scenarios.

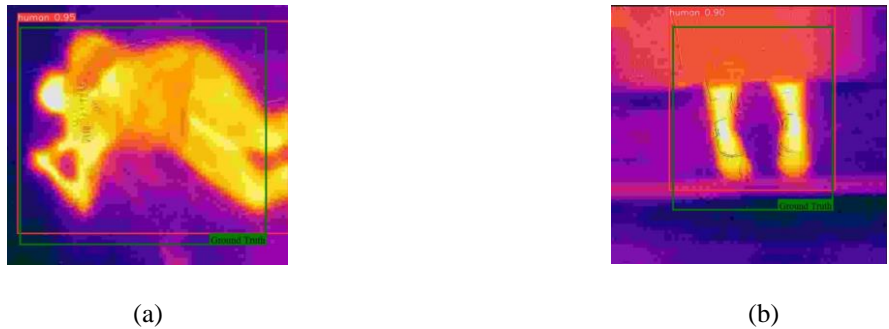

(a)            (b)

*Figure 9. Results of survivor detection algorithm (a) full body visible; (b) legs are exposed while body is covered.*

## 4. Conclusions

This paper has presented and validated an autonomous MAV-supported post-disaster inspection pipeline for structural damage inspection and survivor detection in the building indoor scenario. The proposed methodology incorporates autonomous path planning and navigation, and advanced image and point cloud processing techniques. Experimental studies have demonstrated that the proposed method can be effectively used to detect structural damage and human survivors with high accuracy.

In addition, limitations and future studies are briefly summarized. First, this study employs a monocular camera for image collection, where the reconstructed 3D point cloud is up to scale. Additional calibration of the 3D point cloud to the real-world physical dimension is required. As a future study, a more advanced stereo camera will be mounted to the MAV to achieve 3D reconstruction at the real-world scale without post-calibration. Second, due to the limitation of the 2D LiDAR sensor, the MAV flight height is chosen as a fixed number in the present study. Future research should explore the possibility of a more advanced 3D LiDAR or stereo vision-based SLAM approach for 3D mapping, which will allow the MAV to change flight height as needed during navigation. This will enhance the MAV's capability to pass through narrow openings and avoid obstacles, which is crucial in the complex post-disaster scenario full of fallen debris and objects.

## 5. Acknowledgements

The authors would like to acknowledge the funding and support provided by The Hong Kong University of Science and Technology, Natural Sciences and Engineering Research Council of Canada (NSERC), Any opinions, findings, conclusions, or recommendations expressed in this paper are those of the authors.

## 6. References

[1] Teskeredzic E, Akagic A. Low cost UGV platform for autonomous 2D navigation and map-building based on a single sensory input. In: 2020 7th International Conference on Control, Decision and Information Technologies (CoDIT); 2020; 1:988–93.

[2] Wartman J, Berman JW, Bostrom A, Miles S, Olsen M, Gurley K, Irish J, Lowes L, Tanner T, Dafni J, Grilliot M, Lyda A, Peltier J. Research Needs, Challenges, and Strategic Approaches for Natural Hazards and Disaster Reconnaissance. Frontiers in Built Environment. 2020;6. Available from: https://www.frontiersin.org/articles/10.3389/fbuil.2020.573068

[3] Zaheer Aziz M, Mertsching B. Survivor search with autonomous UGVs using multimodal overt attention. In: 2010 IEEE Safety Security and Rescue Robotics; 2010; p. 1–6. Available from: https://doi.org/10.1109/SSRR.2010.5981566

[4] Thrun S. Learning metric-topological maps for indoor mobile robot navigation. Artificial Intelligence. 1998;99(1):21–71. Available from: https://doi.org/10.1016/S0004-3702(97)00078-7

[5] Yamauchi B. Autonomous urban reconnaissance using man-portable UGVs. In: Unmanned Systems Technology VIII; 2006;6230:261–71. Available from: https://doi.org/10.1117/12.660435

[6] Harik EHC, Guérin F, Guinand F, Brethé J-F, Pelvillain H. UAV-UGV cooperation for objects transportation in an industrial area. In: 2015 IEEE International Conference on Industrial Technology (ICIT); 2015; p. 547–52. Available from: https://doi.org/10.1109/ICIT.2015.7125156

[7] Kim JH, Kwon J-W, Seo J. Multi-UAV-based stereo vision system without GPS for ground obstacle mapping to assist path planning of UGV. Electronics Letters. 2014;50(20):1431–2. Available from: https://doi.org/10.1049/el.2014.2227

[8] Kim CE, Oghaz MM, Fajtl J, Argyriou V, Remagnino P. A Comparison of Embedded Deep Learning Methods for Person Detection. In: VISIGRAPP; 2018. Available from: https://api.semanticscholar.org/CorpusID:54464071

[9] Dong J, Ota K, Dong M. Real-Time Survivor Detection in UAV Thermal Imagery Based on Deep Learning. In: 2020 16th International Conference on Mobility, Sensing and Networking (MSN); 2020; p. 352–9. Available from: https://doi.org/10.1109/MSN50589.2020.00065

[10] Dong J, Ota K, Dong M. UAV-Based Real-Time Survivor Detection System in Post-Disaster Search and Rescue Operations. IEEE Journal on Miniaturization for Air and Space Systems. 2021;2(4):209–19. Available from: https://doi.org/10.1109/JMASS.2021.3083659

[11] Yamane T, Chun P, Dang J, Honda R. Recording of bridge damage areas by 3D integration of multiple images and reduction of the variability in detected results. Computer-Aided Civil and Infrastructure Engineering. 2023;38(17):2391–407. Available from: https://doi.org/10.1111/mice.12971

[12] Pan X, Yang TY. 3D vision-based bolt loosening assessment using photogrammetry, deep neural networks, and 3D point-cloud processing. Journal of Building Engineering. 2023;70:106326. Available from: https://doi.org/10.1016/j.jobe.2023.106326

[13] Ravichandran R, Ghose D, Das K. UAV Based Survivor Search during Floods. In: 2019 International Conference on Unmanned Aircraft Systems (ICUAS); 2019; p. 1407–15. Available from: https://doi.org/10.1109/ICUAS.2019.8798127

[14] Shetty SJ, Ravichandran R, Tony LA, Abhinay NS, Das K, Ghose D. Chapter 16—Implementation of survivor detection strategies using drones. In: Koubaa A, Azar AT, editors. Unmanned Aerial Systems. Academic Press; 2021. p. 417–38. Available from: https://doi.org/10.1016/B978-0-12-820276-0.00023-6

[15] Zhuge S, Xu X, Zhong L, Gan S, Lin B, Yang X, Zhang X. Noncontact deflection measurement for bridge through a multi-UAVs system. Computer-Aided Civil and Infrastructure Engineering. 2022;37(6):746–61. Available from: https://doi.org/10.1111/mice.12771

[16] Meng S, Gao Z, Zhou Y, He B, Djerrad A. Real-time automatic crack detection method based on drone. Computer-Aided Civil and Infrastructure Engineering. 2023;38(7):849–72. Available from: https://doi.org/10.1111/mice.12918

[17] Dowling L, Poblete T, Hook I, Tang H, Tan Y, Glenn W, Unnithan RR. Accurate indoor mapping using an autonomous unmanned aerial vehicle (UAV) [arXiv:1808.01940]. 2018.

[18] Farooq A, Anastasiou A, Souli N, Laoudias C, Kolios PS, Theocharides T. UAV Autonomous Indoor Exploration and Mapping for SAR Missions: Reflections from the ICUAS 2022 Competition. In: 2022 19th International Conference on Ubiquitous Robots (UR); 2022; p. 621–6. Available from: https://doi.org/10.1109/UR55393.2022.9866527

[19] Pan X, Tavasoli S, Yang TY. Autonomous 3D vision-based bolt loosening assessment using micro aerial vehicles. Computer-Aided Civil and Infrastructure Engineering. 2023;38(17):2443–54. Available from: https://doi.org/10.1111/mice.13023

[20] Tavasoli S, Pan X, Yang TY. Real-time autonomous indoor navigation and vision-based damage assessment of reinforced concrete structures using low-cost nano aerial vehicles. Journal of Building Engineering. 2023;68:106193. Available from: https://doi.org/10.1016/j.jobe.2023.106193

[21] Tavasoli S, Pan X, Yang TY, Gazi S, Azimi M. Autonomous damage assessment of structural columns using low-cost micro aerial vehicles and multi-view computer vision. Available from: https://doi.org/10.48550/arXiv.2308.16278

[22] Pliakos C, Vlachos S, Bliamis C, Yakinthos K. Preliminary design of a multirotor UAV for indoor search and rescue applications. Journal of Physics Conference Series. 2024;2716:012067. Available from: https://doi.org/10.1088/1742-6596/2716/1/012067

[23] Tavasoli S, Poorghasem S, Pan X, Yang TY, Bao Y. Autonomous post-disaster indoor navigation and survivor detection using low-cost micro aerial vehicles. Computer-Aided Civil and Infrastructure Engineering.

[24] Pan X. Three-dimensional vision-based structural damage detection and loss estimation–towards more rapid and comprehensive assessment [Doctoral dissertation]. University of British Columbia; 2022. Available from: http://hdl.handle.net/2429/83361

[25] Xiao Y, Pan X, Tavasoli S, Azimi M, Bao Y, Noroozinejad Farsangi E, Yang TY. Autonomous inspection and construction of civil infrastructure using robots. In: Automation in Construction Toward Resilience. Noroozinejad Farsangi E, Noori M, Yang T, Lourenço PB, Gardoni P, Takewaki I, Chatzi E, Li S, editors. 2023. Available from: https://doi.org/10.1201/9781003325246

[26] Pan X, Yang TY. Postdisaster image-based damage detection and repair cost estimation of reinforced concrete buildings using dual convolutional neural networks. Computer-Aided Civil and Infrastructure Engineering. 2020;35:495–510. Available from: https://doi.org/10.1111/mice.12549

[27] Li C, Su R, Pan X. Assessment of out-of-plane structural defects using parallel laser line scanning system. Computer-Aided Civil and Infrastructure Engineering. 2024;39(6):834–51. Available from: https://doi.org/10.1111/mice.13102

[28] Pan X, Yang TY. 3D vision-based out-of-plane displacement quantification for steel plate structures using structure from motion, deep learning, and point cloud processing. Computer-Aided Civil and Infrastructure Engineering. 2023;38:547–61. Available from: https://doi.org/10.1111/mice.12906

[29] Pan X, Yang TY, Xiao Y, Yao H, Adeli H. Vision-based real-time structural vibration measurement through interactive deep-learning-based detection and tracking methods. Engineering Structures. 2023;281:115676. Available from: https://doi.org/10.1016/j.engstruct.2023.115676

[30] Xie F, Pan X, Yang TY, Ernewein B, Li M, Robinson D. A novel computer vision and point cloud-based approach for accurate structural analysis of a tall irregular timber structure. Structures. 2024. Available from: https://doi.org/10.1016/j.istruc.2024.107697

[31] Pan X, Yang TY. Image-based monitoring of bolt loosening through deep-learning-based integrated detection and tracking. Computer-Aided Civil and Infrastructure Engineering. 2022;37(10):1207–22. Available from: https://doi.org/10.1111/mice.12797

[32] Pan X, Yang TY. Bolt loosening assessment using ensemble vision models for automatic localization and feature extraction with target-free perspective adaptation. Computer-Aided Civil and Infrastructure Engineering. 2024;1-16. Available from: https://doi.org/10.1111/mice.13355

[33] Pan X, Yang TY, Liu R, Xiao Y, Xie F. A computer vision and point cloud-based monitoring approach for automated construction tasks using full-scale robotized mobile cranes. Journal of Intelligent Construction. 2024. Available from: https://doi.org/10.26599/JIC.2025.9180086

[34] Yamauchi B. A frontier-based approach for autonomous exploration. In: Proceedings 1997 IEEE International Symposium on Computational Intelligence in Robotics and Automation CIRA'97; 1997. p. 146–51. Available from: https://doi.org/10.1109/CIRA.1997.613851